\documentclass{article}
\usepackage{spconf,amsmath,graphicx,amssymb,comment,bm,hyperref, bbm}

\title{PEFT for Speech: Unveiling Optimal Placement, Merging Strategies, and Ensemble Techniques}

\name{Tzu-Han Lin$^\dagger$, How-Shing Wang$^\dagger$, Hao-Yung Weng$^\ddagger$, Kuang-Chen Peng$^\ddagger$, Zih-Ching Chen$^*$, Hung-yi Lee$^*$\thanks{$^{\dagger \ddagger}$Equal contribution. $^*$Corresponding Author}}
\address{National Taiwan University}

\begin{document}

\maketitle

\begin{abstract}

Parameter-Efficient Fine-Tuning (PEFT) is increasingly recognized as an effective method in speech processing. However, the optimal approach and the placement of PEFT methods remain inconclusive. Our study conducts extensive experiments to compare different PEFT methods and their layer-wise placement adapting Differentiable Architecture Search (DARTS). We also explore the use of ensemble learning to leverage diverse PEFT strategies. The results reveal that DARTS does not outperform the baseline approach, which involves inserting the same PEFT method into all layers of a Self-Supervised Learning (SSL) model. In contrast, an ensemble learning approach, particularly one employing majority voting, demonstrates superior performance. Our statistical evidence indicates that different PEFT methods learn in varied ways. This variation might explain why the synergistic integration of various PEFT methods through ensemble learning can harness their unique learning capabilities more effectively compared to individual layer-wise optimization.

\end{abstract}


\section{Introduction}
\label{sec:intro}

In recent years, SSL models have demonstrated remarkable improvements in a variety of downstream tasks in the speech domain \cite{hsu2021hubert,  radford2022whisper, mohamed2022self}.  However, fine-tuning these models is computationally expensive, which brings into focus the importance of PEFT \cite{ding2022delta, houlsby2019parameter, fu2022adapterbias, kessler2022adapter, pfeiffer2020adapterfusion}. PEFT, which includes methods such as Houlsby adapters\cite{houlsby2019parameter} and Low Rank Adaptation (LoRA) \cite{hu2021lora}, offers a promising alternative to traditional fine-tuning techniques by reducing computational and storage overhead while maintaining, or even improving, performance of fine-tuning SSL speech models \cite{chen2023parameter, chen2023exploring, chang2022exploration}.

Since previous studies suggest that various layers of an SSL model may capture different aspects of information 
\cite{pasad2021layer, chen2022chapter, chen2023exploring}, the placement of adapters within the pre-trained model may be crucial.  While various studies have explored different methods to determine the optimal placement of PEFT methods in Natural Language Processing (NLP) \cite{moosavi2022adaptable,  zhou2023autopeft}, there has been little prior investigation in the speech domain \cite{findadapnet}. In light of this, we explore various methods, including identifying the optimal placement, combining various PEFT methods, and employing ensemble learning, for speech tasks, including Automatic Speech Recognition (ASR), Phoneme Recognition (PR), Speaker Identification (SID), Speaker Diarization (SD), Slot Filling (SF), and Emotion Recognition (ER).

In this work, we leveraged DARTS \cite{liu2018darts} to identify the optimal placement of PEFT methods. Additionally, as observed in \cite{hydra}, which notes that different adapters may capture different aspect of information in NLP tasks, we conduct experiments on merging multiple PEFT methods. Lastly, in contrast to jointly training, where each PEFT module operates dependently, we explore different ensemble learning strategies.

Our main contributions are: \begin{enumerate}
\item We conducted extensive experiments to compare different PEFT methods and combined them with techniques like ensemble learning and DARTS, which is the first time such methods have been introduced to the speech processing field for PEFT method selection.
\item While architecture search methods have achieved success in NLP, our exploration revealed that optimizing PEFT method selection with DARTS does not outperform the straightforward approach of inserting a single PEFT module into a pre-trained model.
\item We found that ensembling different PEFT method outputs with majority voting, under the same parameter amount constraint, yields better results than using a single PEFT method.
\end{enumerate}

\section{Method}
\label{sec:method}

In our study on optimizing PEFT methods for SSL speech models, we examine three approaches. Firstly, in subsection~\ref{subsec: darts} we employ DARTS \cite{liu2018darts} to strategically place the different PEFT methods within transformer layers. Secondly, subsection~\ref{subsec: hybrid} merges different PEFT methods within each transformer layer. Finally, in subsection~\ref{subsec: ensemble}, we investigate the effectiveness of ensembling different PEFT methods. The PEFT methods explored in this study include sequential and parallel Houlsby adapters \cite{houlsby2019parameter}, as well as LoRA \cite{hu2021lora}.

\subsection{Layer-Wise Optimization of PEFT Selection}
\label{subsec: darts}

While previous studies have explored the search for the optimal structure of PEFT methods \cite{moosavi2022adaptable}, few of them have investigated the efficacy of such searches on speech processing tasks. We leveraged DARTS to find the optimal layer-wise placement of PEFT methods for speech processing tasks.

DARTS is a gradient-based Network Architecture Search (NAS) method \cite{ren2021comprehensive}. Different from the original DARTS, which conducts a cell-based search process, our approach focuses on determining the optimal placement of PEFT methods within each transformer layer. To the best of our knowledge, this is the first attempt to utilize DARTS for the placement of PEFT methods within SSL models in the speech domain.

The training process is divided into two stages: architecture search and network training. In the first stage, DARTS is used to determine the best architecture, utilizing two halves of the training set separately for architecture and network weights. The optimization objective is aligned with the original DARTS. To reduce computational costs, we omit the second derivative term in the gradient of the validation loss during the search, akin to the first-order MAML \cite{nichol2018first}.

Let $S^{(i)} = [\mathcal{A}^{(i)}_1, \mathcal{A}^{(i)}_2, \ldots, \mathcal{A}^{(i)}_N]$ represent the candidate PEFT methods for layer $i$, $\bm{\alpha}^{(i)} \in \mathbb{R}^N$ denote the weight vector for each module in layer $i$, and $\mathbf{x}^{(i)}$ denote the input representation of layer $i$. The output representation of layer $i$, generated by employing module $\mathcal{A}^{(i)}_n$, is denoted by $\mathcal{A}^{(i)}_n(\mathbf{x}^{(i)})$. The layer $i$ latent representation $\mathbf{h}^{(i)}$ is obtained by applying softmax to all module outputs: \begin{align*}
    \mathbf{h}^{(i)} = \sum^{N}_{n=1} \frac{\exp{(\alpha^{(i)}_{n})}}{\sum^{N}_{n'=1} \exp{(\alpha^{(i)}_{n'})}} \mathcal{A}^{(i)}_n(\mathbf{x}^{(i)})
\end{align*}

In the second stage, we select the module with the highest weight as the final choice for each layer. i.e. $\mathcal{A}^{(i)} = \mathcal{A}^{(i)}_n$, where $n = \arg\max_{n'} \alpha^{(i)}_{n'}$. Once the architecture is determined, we proceed to refine the network weights using the entire training set.

\subsection{Hybrid Method}
\label{subsec: hybrid}
\begin{figure}[htb]

\begin{minipage}[b]{1.0\linewidth}
  \centering
  \centerline{\includegraphics[width=8.5cm]{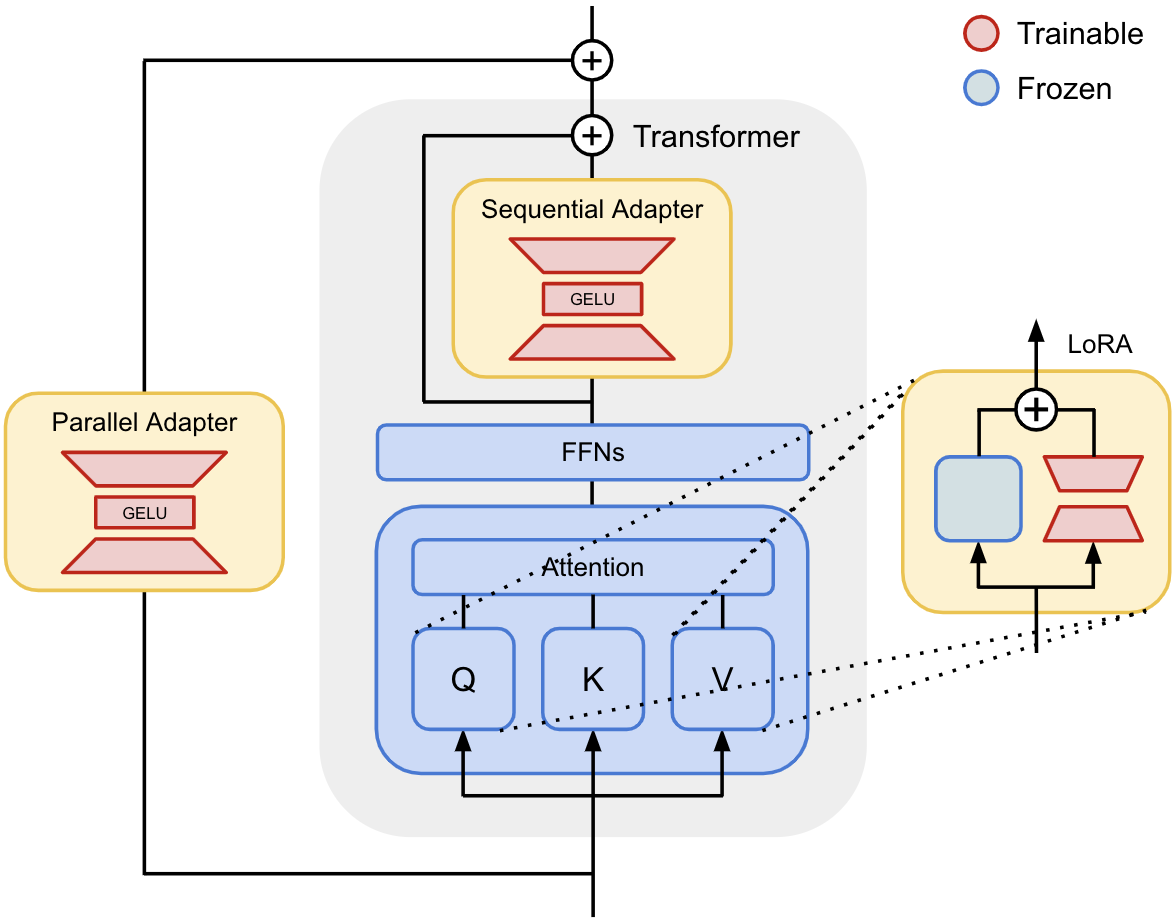}}
\end{minipage}

\caption{The architecture of the Hybrid Method. The trainable/frozen parameters are colored in red/blue.}
\label{fig:hybrid}

\end{figure}

In contrast to the intricate DARTS method, we adopt a simpler approach by merging the PEFT methods within each transformer layer, as illustrated in Figure~\ref{fig:hybrid}. The original framework was proposed by \cite{hydra}. Building on the observation that sequential and parallel Houlsby adapters may capture various aspects of information, they introduced both into each transformer layer. In this section, we extend this framework to merge multiple PEFT methods within each transformer layer.

\subsection{Ensemble Learning}
\label{subsec: ensemble} 
To further explore a combination of approaches of different PEFT methods under a parameter size constraint, we investigated ensemble learning on outputs from models trained with different PEFT methods. We adopted simple ensembling approaches such as Majority Voting or Averaging Output Probabilities, and subsequently applied sequence alignment to boost the performance of the aforementioned approaches on tasks that use CTC \cite{graves2006connectionist} loss in training.

\subsubsection{Majority Voting}
\label{subsec:majvote}

Let $\mathbf{P}_1, \mathbf{P}_2, \dots, \mathbf{P}_N$ represent the probability outputs of $N$ models, where $\mathbf{P}_i = [p^{(i)}_{1}, p^{(i)}_{2}, \dots, p^{(i)}_{C}]$ denotes the predicted probabilities for $C$ classes by the $i^{th}$ model. For a input instance, the predictions from each model are aggregated. Define $I_{ik}=\mathbbm{1}(\arg\max_j p^{(i)}_{j}=k)$, where $\mathbbm{1}(\cdot)$ is the indicator function. Additionally, let $\mathbf{V} = [v_1, v_2, \dots, v_C]$ represent the voting vector, where $v_j = \sum_{i=1}^{N}I_{ij}$ is the count of votes for class $j$. The class with the highest count in the voting vector $\mathbf{V}$ is selected as the final prediction $\hat{y}$. i.e. 
$\hat{y} = \arg\max_{j} v_j$.

\subsubsection{Average Output Logits}
\label{subsec:avgoutput}
 In this approach, the average probability for each class is computed across the $N$ models. The average probability $\bar{p}_j$ for class $j$ is calculated as: $\bar{p}_j = \frac{1}{N} \sum_{i=1}^{N} p^{(i)}_{j}$, where $p^{(i)}_{j}$ is the probability predicted by the $i$-th model for class $j$. The output decision is $\hat{y} = \arg\max_{j} \bar{p}_j$.

\subsubsection{CTC: Alignment Study}

CTC Loss is widely adopted in sequence-to-sequence tasks. However, the alignments of the output sequence with the input sequence may differ across different models. For example, the same positions in the output sequence from $N$ different models may correspond to different positions in the input sequence. Therefore, simply averaging or voting on the token distributions generated by each model may not yield the desired results. As shown in Table~\ref{tab: fullresults}, tasks that use CTC loss such as PR and SF, showed performance degradation under our aforementioned ensemble approaches. Addressing this alignment issue, we modified the progressive alignment approach for Multi-Sequqnce-Alignment (MSA) from \cite{praveen2021warped}, which uses Dynamic Time Warping (DTW) to align output sequences of $N$ different models. Let $S_i=[\mathbf{s}^{(i)}_{1}, \mathbf{s}^{(i)}_{2}, ..., \mathbf{s}^{(i)}_{T}]$ be the output sequence of the $i^{th}$ model, where $T$ is the sequence length for all $N$ models and $\mathbf{s}^{(i)}_{t}$, the $t^{th}$ element of $S_i$, is a token-distribution. After MSA, we obtain N aligned sequences, where $\bar{S}_i=[\mathbf{s}^{(i)}_{1}, \mathbf{s}^{(i)}_{2}, ..., \mathbf{s}^{(i)}_{\bar{T}}]$ and $\bar{T} \ge T$.  We didn't adopt the Blank Removal feature in \cite{praveen2021warped} since it didn't improve the performance. For the obtained aligned sequences, we performed the averaging method mentioned in subsection \ref{subsec:avgoutput}, which is the uniform weights version of \cite{praveen2021warped}, as well as applied the voting approach in subsection \ref{subsec:majvote}.

\section{Experiment}
\label{sec:experiment}

\subsection{SUPERB benchmark}
\label{subsec:SUPERB benchmark}

We selected 6 tasks across 4 different aspects from the SUPERB benchmark \cite{yang2021superb} for evaluation, which includes Automatic Speech Recognition (ASR), Phoneme Recognition (PR), Speaker Identification (SID), Speaker Diarization (SD), Slot Filling (SF), and Emotion Recognition (ER). In addition to the task-specific metrics, we introduce a metric to aggregate the task-specific scores into a single score\footnote{The original measure can be found in \url{https://superbbenchmark.org/challenge-slt2022/metrics}.}. Since our experiments are conducted on HuBERT, we map the performance of fine-tuned HuBERT to 1000, instead of the SoTA on the task, for better comparison of each method.

Additionally, MiniSUPERB \cite{wang2023minisuperb} suggests that the performance of PR and SID is strongly correlated with the final performance on the SUPERB benchmark. Consequently, we initially conducted our experiments on PR and SID to estimate the overall performance of each method.

\subsection{Experiment Setup}
In our experiment, we use HuBERT \cite{hsu2021hubert} as the upstream model. For the baseline methods, in addition to incorporating the PEFT methods described earlier, we also include the full fine-tuning and weighted-sum methods from \cite{yang2021superb}. In the case of the weighted-sum method, HuBERT is frozen, and only the weights associated with each layer are tuned. For the PEFT methods' configuration, we set the bottleneck dimension of Houlsby adapters to 32, and the rank of LoRA is set to 8. Additionally, the weighted-sum method is included in the training of PEFT methods. The best learning rate is found by searching in a range between $1 \times 10^{-6}$ and $1 \times 10^{-2}$.

Additionally, to maintain a similar number of trainable parameters in the upstream model across all methods during inference time, we set the bottleneck dimension of Houlsby adapters to 10 and the rank of LoRA to 2 for the Hybrid method and ensemble learning. This adjustment results in the number of parameters for each PEFT module being strictly less than one-third of the original setup.

As for the DARTS method, we set the number of steps in stage-1 to 25\% of the total steps. We denote this setup as 'DARTS 25\%.' Additionally, recognizing that training only the derived model from stage-1 on the full training set for the remaining steps, instead of the full training steps, may potentially lead to performance degradation, we include an extra experiment. In this experiment, we initialize the network architecture with the result from stage-1 of DARTS training and then train this re-initialized network for the full number of training steps. This setup is referred to as 'DARTS retrain.'

\subsection{Results}

Due to budget constraints, we initially conducted experiments on PR and SID, as the performance of these two tasks strongly correlates with the final performance on the SUPERB benchmark \cite{wang2023minisuperb}. The results are presented in Table~\ref{tab: minisuperb}. Among the baselines (rows (a)--(c)), inserting parallel Houlsby adapters into all layers of the SSL model yields the best result, which outperforms both DARTS-related methods (rows (d)--(e)) despite their high computation costs. The remaining methods yield comparable results. Consequently, we exclude DARTS methods from subsequent evaluations.

The final results across six different tasks are presented in Table~\ref{tab: fullresults}. All methods demonstrate improvements compared to full fine-tuning (row (a)). With the exception of the 'Avg Logits' (row (g)), the remaining methods all surpass the weighted-sum (row (b)). Notably, Voting (row (h)) outperforms other methods, achieving the highest SUPERB score. Interestingly, the Hybrid method (row (f)) fails to outperform the single PEFT module insertion methods (rows (a)--(c)). Lastly, incorporating sequence alignment before the averaging and voting show improvements in tasks that adopt CTC loss.

\begin{table*}[ht!]
\renewcommand\thetable{2}
\renewcommand{\arraystretch}{1.2}
\centering
\setlength{\tabcolsep}{1.5mm}{
\scalebox{0.8}[0.8]{
\begin{tabular}{@{}ll|c|cc|cc|c|c|c@{}}
 & Method & \# Params & ASR & PR & SID & SD & SF & ER & SUPERB Score \\
 & & & WER $\downarrow$ & PER $\downarrow$ & Acc $\uparrow$ & DER $\downarrow$ & F1 $\uparrow$ & Acc $\uparrow$ & $superb_s$ $\uparrow$ \\
\hline
(a) & Fine-tune & 94.7M & 6.35 & \textbf{2.45} & 64.56 & 9.32 & 86.17 & \textbf{69.95} & 1000 \\
(b) & Weighted Sum & 12 & 6.42 & 5.41 & 81.42 & 5.88 & 86.71 & 64.92 & 1808 \\
\hline
(c) & Sequential & 0.6M & 6.73 & 2.66 & 90.35 & 4.38 & 86.38 & 60.93 & 2153 \\
(d) & Parallel & 0.6M & 5.55 & 2.51 & 92.24 & 4.41 & 85.17 & 59.28 & 2140 \\
(e) & LoRA & 0.29M & 5.53 & 3.15 & 90.95 & 5.26 & 86.75 & 62.41 & 1980 \\
\hline
(f) & Hybrid & 0.46M & 5.56 & 2.63 & 92.87 & 4.50 & 84.86 & 58.18 & 2112 \\
\hline
(g) & Avg Logits & 0.46M & \textbf{5.20} / 5.25 & 2.68 / 3.84 & \textbf{94.14} & 7.81 & \textbf{88.26} / 85.02 & 62.85 & 1433 / 1397 \\
(h) & Voting & 0.46M & 5.26 / 5.38 & 2.71 / 3.26 & 92.52 & \textbf{4.24} & 87.36 / 85.66 & 63.68 & \textbf{2239} / 2219 \\
\hline
\end{tabular}
}
}

\caption{Results of different methods. The second column indicates the additional trainable parameters used in the upstream model. The numbers preceding the "/" indicate that we applied alignment before the avg/voting operation.}

\label{tab: fullresults}
\end{table*}

\begin{table}
\renewcommand\thetable{1}
\renewcommand{\arraystretch}{1.2}
\centering
\setlength{\tabcolsep}{1.5mm}{
\scalebox{0.75}[0.75]{
\begin{tabular}{@{}ll|c|c|c@{}}
 & Method & PR & SID & SUPERB Score \\
 & & PER $\downarrow$ & Acc $\uparrow$ & $superb_s$ $\uparrow$ \\
\hline
(a) & Sequential & 2.66 & 90.35 & 1288 \\
(b) & Parallel & \textbf{2.51} & 92.24 & 1311 \\
(c) & LoRA & 3.15 & 90.95 & 1292 \\
\hline
(d) & DARTS 25\% & 2.63 & 90.58 & 1291 \\
(e) & DARTS retrain & 2.59 & 90.75 & 1293 \\
\hline
(f) & Hybrid & 2.63 & 92.87 & 1317 \\
\hline
(g) & Avg Logits & 2.68 / 3.84 & \textbf{94.14} & \textbf{1331} / 1324 \\
(i) & Voting & 2.71 / 3.26 & 92.52 & 1313 / 1309 \\
\hline
\end{tabular}
}
}

\caption{Performance of PR and SID}
\label{tab: minisuperb}
\end{table}

\section{Discussion}
\label{sec:discussion}

\begin{figure}[htb]

\begin{minipage}[b]{1.0\linewidth}
  \centering
  \centerline{\includegraphics[width=8.5cm]{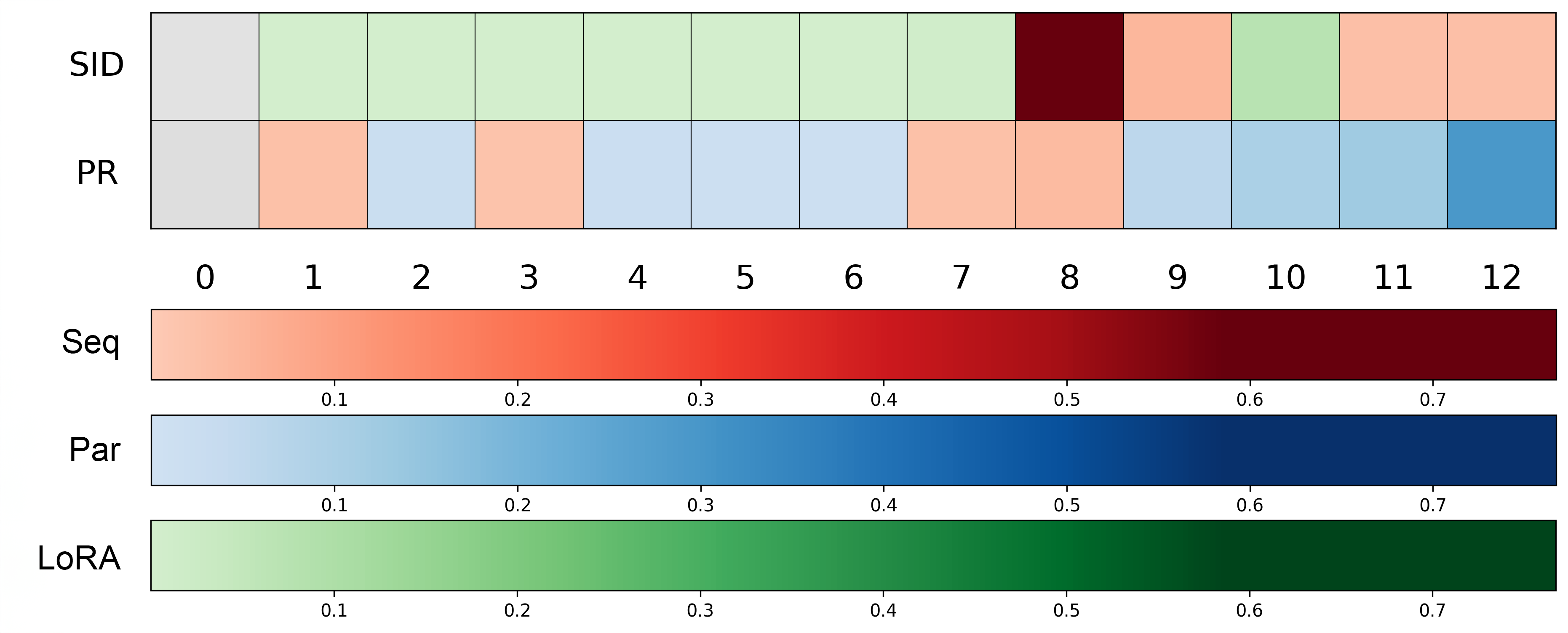}}
\end{minipage}

\caption{Selected PEFT methods for each layer and their associated weights for PR and SID. The shade intensity of each cell indicates the weight associated with each layer.}
\label{fig:adapter_weight}

\end{figure}

\noindent\textbf{Layer-Wise Optimization of PEFT Selection}: The architecture derived from DARTS for PR and SID differs significantly, as shown in Figure~\ref{fig:adapter_weight}, indicating its potential for bespoke PEFT method optimization. However, as shown in Table~\ref{tab: minisuperb}, the performance (rows (d)--(e)) fails to surpass the baselines (rows (a)--(c)). During the experiment, we observed that DARTS's performance is highly sensitive to the number of steps used for architectural search, aligning with prior research \cite{1909.09656}. In our experimental setup, 25\% of the total steps were allocated to architectural search, which might not have been optimal for this model or task, resulting in poor performance. We also speculate that the optimal number of steps for architectural search might vary from task to task and model to model. A more thorough exploration of DARTS in various contexts is planned for future work.

\noindent\textbf{Hybrid Method}: As shown in Table~\ref{tab: fullresults}, the result of the Hybrid method (row (f)) did not surpass that of the single PEFT methods (rows (a)--(c)). Since PEFT methods do not operate independently and might influence the learning of others, we suspect that this complexity contributes to its failure to outperform the single PEFT module insertion.

\noindent\textbf{Ensemble Learning}: Due to the failure of the Hybrid method, we adopted ensemble learning strategies to explore whether combining the outputs of each PEFT module independently could improve performance. As shown in Table~\ref{tab: fullresults}, ensemble methods (rows (g)--(h)) indeed improved the overall performance. The worsened DER on SD (row(g)) is under investigation. Moreover, the searched architectures for PR and SID differ, as shown in Figure~\ref{fig:adapter_weight}. These results suggest that different PEFT methods may capture different information. To dig into this, we performed statistical tests on predictions generated by PEFT methods used in ensemble. For ASR and PR, we conducted the MAPSSWE test. For SID and ER, the McNemar test was applied. As for SF and SD, we adopted the Student's t-test. The resulting p-values are reported in Table~\ref{tab: stats}. Significantly, except for ASR, differences between the Houlsby adapter and LoRA were observed. Moreover, distinctions between sequential and parallel Houlsby adapters were evident in SID and ER. These findings substantiate our hypothesis that different PEFT methods may capture distinct aspects of information across various tasks.

\begin{table}
\renewcommand{\arraystretch}{1.2}
\centering
\setlength{\tabcolsep}{1.5mm}
{
\scalebox{0.8}{
\begin{tabular}{@{}l|cccccc@{}}
 & ASR & PR & SID & ER & SF & SD\\
\hline
 Seq vs Par & \textbf{0.741} & \textbf{1.000} & 0 & 0 & \textbf{0.3230} & \textbf{0.1826} \\
 Seq vs LoRA & \textbf{0.741} & 0 & 0 & 0 & 0 & 0 \\
 Par vs LoRA & \textbf{0.497} & 0 & 0 & 0 & 0 & 0 \\
\hline
\end{tabular}
}
}
\caption{p-value comparison of PEFT methods per task, with "Seq" for sequential and "Par" for parallel. Bold cells indicate cases where the difference is insignificant. (p-value $>$ 0.05)}
\label{tab: stats}
\end{table}

\section{Conclusion}
\label{sec:conclusion}

Our results demonstrate that the ensemble learning approach, particularly when employing a voting mechanism, yields the best performance. This outcome contrasts with the performance of DARTS, which, despite its potential for bespoke PEFT method optimization, did not surpass the baseline results. These findings suggest that different PEFT methods may possess diverse learning capabilities, which can be more effectively exploited through a synergistic ensemble approach rather than through individualized layer-wise optimization. We support this conclusion with statistical evidence, highlighting the potential of ensemble learning in enhancing speech processing tasks with diverse PEFT strategies.

\vfill\pagebreak

\small\bibliographystyle{IEEEbib}
\bibliography{refs}

\end{document}